\documentclass{article} 
\usepackage{iclr2025_conference,times}

\usepackage{amsmath,amsfonts,bm}

\def\eqref#1{equation~\ref{#1}}

\def\1{\bm{1}}

\DeclareMathAlphabet{\mathsfit}{\encodingdefault}{\sfdefault}{m}{sl}
\SetMathAlphabet{\mathsfit}{bold}{\encodingdefault}{\sfdefault}{bx}{n}

\usepackage{hyperref}
\usepackage{url}

\usepackage{amsmath}
\usepackage{amsfonts}
\usepackage{graphicx}
\usepackage{algorithm}
\usepackage{algorithmic}

\usepackage{tcolorbox}
\usepackage{booktabs}
\usepackage{wrapfig}

\title{TradExpert: Revolutionizing Trading with Mixture of Expert LLMs}

\author{Qianggang Ding$^{1,2}$, Haochen Shi$^{1,2}$, Jiadong Guo$^3$, Bang Liu$^{1,2}$\thanks{Corresponding author. Canada CIFAR AI Chair.} \\
$^1$ Université de Montréal, $^2$ Mila, $^3$ The Hong Kong University of Science and Technology \\
\texttt{\{qianggang.ding, haochen.shi\}@umontreal.ca, jguobc@connect.ust.hk} \\
\texttt{bang.liu@umontreal.ca}
}

\iclrfinalcopy 
\begin{document}

\maketitle

\begin{abstract}
The integration of Artificial Intelligence (AI) in the financial domain has opened new avenues for quantitative trading, particularly through the use of Large Language Models (LLMs). However, the challenge of effectively synthesizing insights from diverse data sources and integrating structured and unstructured data persists. This paper presents TradExpert, a novel framework that employs a mix of experts (MoE) approach, using four specialized LLMs, each analyzing distinct sources of financial data, including news articles, market data, alpha factors, and fundamental data. The insights of these expert LLMs are further synthesized by a General Expert LLM to make a final prediction or decision. With specific prompts, TradExpert can be switched between the prediction mode and the ranking mode for stock movement prediction and quantitative stock trading, respectively. In addition to existing benchmarks, we also release a large-scale financial dataset to comprehensively evaluate TradExpert's effectiveness. Our experimental results demonstrate TradExpert's superior performance across all trading scenarios.
\end{abstract}

\section{Introduction}

The fusion of artificial intelligence with financial analytics has spawned a new era of innovation, particularly with the infusion of Large Language Models (LLMs) into the realm of finance. These models, which have formerly excelled in natural language processing (NLP) tasks, are now being tailored to decode the complex and cryptic narratives of financial data. This adaptation is driven by a crucial insight: Financial markets are not just numbers-crunching engines but complicated information systems where the subtleties of news articles, reports, and economic indicators interweave to influence market dynamics.

Before the advent of LLMs, traditional financial models~\cite{zeng2023financial,yang2020deep,liu2020finrl,baek2018modaugnet},  primarily relied on quantitative methods such as statistical analysis, time series forecasting, and econometric models. These models often struggled to incorporate unstructured data such as news articles or financial reports without manual intervention. 
As a result, the development of LLMs tailored for financial applications has progressed rapidly. Initial ventures into this domain repurposed general LLMs such as  GPTs~\cite{brown2020language,achiam2023gpt} and LLaMAs~\cite{touvron2023llama1,touvron2023llama,llama3modelcard} to interpret financial texts.
However, more specialized language models such as FinBERT~\cite{araci2019finbert}, BloombergGPT~\cite{wu2023bloomberggpt}, and FinGPT~\cite{yang2023fingpt} have since evolved, demonstrating enhanced proficiency in understanding and predicting market movements from unstructured data. These models were specifically fine-tuned or pre-trained on vast amounts of financial corpus. This extensive training on domain-specific datasets has allowed them to better capture typical patterns in the financial corpus. Despite these advancements, the challenge remains to effectively synthesize insights from diverse data sources like historical stock prices, alpha factors, fundamental data, news articles, etc. In addition, integration of the deluge of unstructured financial texts with structured quantitative metrics still remains to be investigated with language models.

\begin{wrapfigure}{r}{0.5\textwidth}
  \centering
  \includegraphics[width=1.0\linewidth]{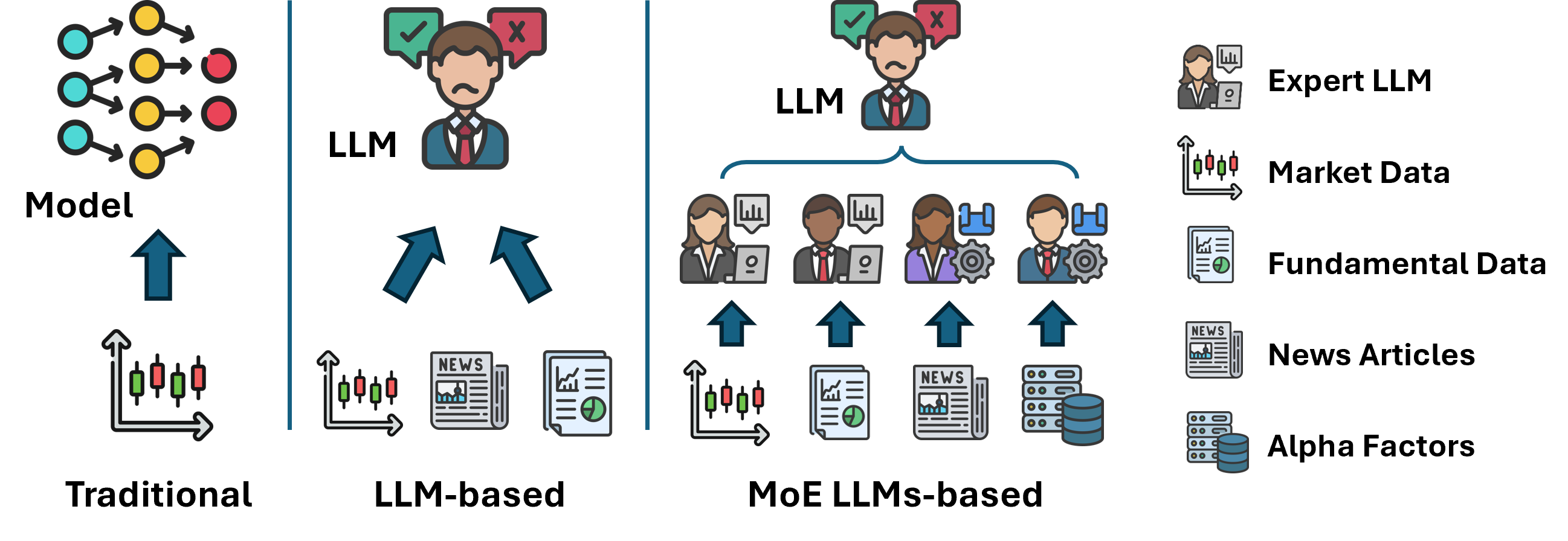}
  \caption{Illustration of traditional, LLM-based, and MoE LLMs-based financial models with diverse financial data sources. }
  \label{fig:intro}
\end{wrapfigure}

To this end, we propose the TradExpert framework, which stands at the confluence of these challenges. It leverages a Mixture of Experts (MoE) approach~\cite{eigen2013learning,du2022glam,shen2023mixture}, involving multiple LLMs each specialized in distinct facets of financial data—news articles, market data, alpha factors, and fundamental data. This not only enhances the model's ability to process diverse data modalities but also allows for a more nuanced understanding of how different factors interact to influence market trends. Figure~\ref{fig:intro} illustrates differences among traditional, LLM-based, and MoE LLMs-based financial models. In TradExpert, each expert works with a distinct focus and produces specialized reports, which are finally summarized and analyzed by a general expert, just like the structured division of labor seen in the real world. Specifically, TradExpert employs specialized LLMs to first independently analyze different data sources, then integrates these analyses via another LLM that synthesizes insights to predict market movements and inform trading strategies. Innovatively, we utilize a reprogramming mechanism to convert time series data to embeddings aligned with LLMs. In addition, we propose two modes for the General Expert LLM, prediction mode and ranking mode, for stock movement prediction and stock trading strategies, respectively. In ranking mode, we innovatively let the LLM serve as a comparator within a relaxed sorting algorithm, enabling the selection of the Top-K ranked stocks for trading. To comprehensively evaluate our method, we have also collected a large-scale datasets, which will be publicly released.

Oue contributions can be summarized as follows.

\begin{itemize}
    \item We present TradExpert, a novel framework that employs an MoE approach, integrating four LLMs each specialized to analyze distinct sources of financial data, imitating the structured division of labors seen in the real world.
    \item We utilize the LLM as a comparator within a relaxed sorting algorithm, which enables trades with the Top-K ranked stocks based on TradExpert's prediction.
    \item We release a comprehensive dataset encompassing a wide range of financial data, which serves as a new benchmark for financial analysis.
    \item Our comprehensive experiments show that TradExpert consistently outperforms state-of-the-art baselines across all trading scenarios. Ablation studies validate the effectiveness of the modules proposed in TradExpert.
\end{itemize}

\section{Related Work}

\paragraph{Financial Language Models} have significantly advanced in recent years, blending NLP techniques with financial analytics to extract meaningful insights from vast amounts of unstructured financial data. To begin with, FinBert~\cite{araci2019finbert} is a financial domain-specific variant of BERT, pretrained on a large corpus of financial communications. In 2023, BloombergGPT~\cite{wu2023bloomberggpt} emerged as a 50-billion-parameter model trained on a vast financial dataset. FLANG~\cite{shah-etal-2022-flang} introduced a financial language model with specialized masking and objectives. Astock~\cite{zou2022astock} provided a platform for studying NLP-aided stock auto-trading algorithms on the Chinese market. BBT-FinT5~\cite{lu2023bbt} advanced Chinese financial NLP with a large-scale pre-trained model. FinMA~\cite{xie2023pixiu} showcased a model fine-tuned on a multi-task instruction datasets. FinGPT~\cite{yang2023fingpt} provided an open-source framework for financial LLMs. InvestLM~\cite{yang2023investlm} showed the effectiveness of instruction tuning for investment-related tasks. FinReport~\cite{li2024finreport} introduced a system for automatic financial report generation. Lastly, AlphaFin~\cite{li2024alphafin} integrated retrieval-augmented generation techniques for financial analysis. Collectively, these works demonstrate the evolution of financial NLP models and benchmarks, advancing the capabilities of LLMs in financial applications. 

\paragraph{Integration of Text and Financial Data} has also been rapidly developed for stock movement prediction. StockNet~\cite{xu2018stock} developed a deep generative model that jointly exploits text and price signals for stock movement prediction. SLOT~\cite{soun2022accurate} improved upon this by using self-supervised learning to handle sparse and noisy tweet data, capturing multi-level price trends. CH-RNN~\cite{wu2018hybrid} introduced a hybrid deep sequential modeling approach that leverages social text for stock prediction, incorporating cross-modal attention mechanisms. More recently, studies~\cite{lopez2023can,chen2023chatgpt} have explored the use of ChatGPT for stock movement prediction, comparing its performance with traditional state-of-the-art models. These works collectively demonstrate the increasing sophistication of models that integrate text and financial data, highlighting the potential for improving trading scenarios.

\begin{figure*}[t]
\centering
\includegraphics[width=1.0\textwidth]{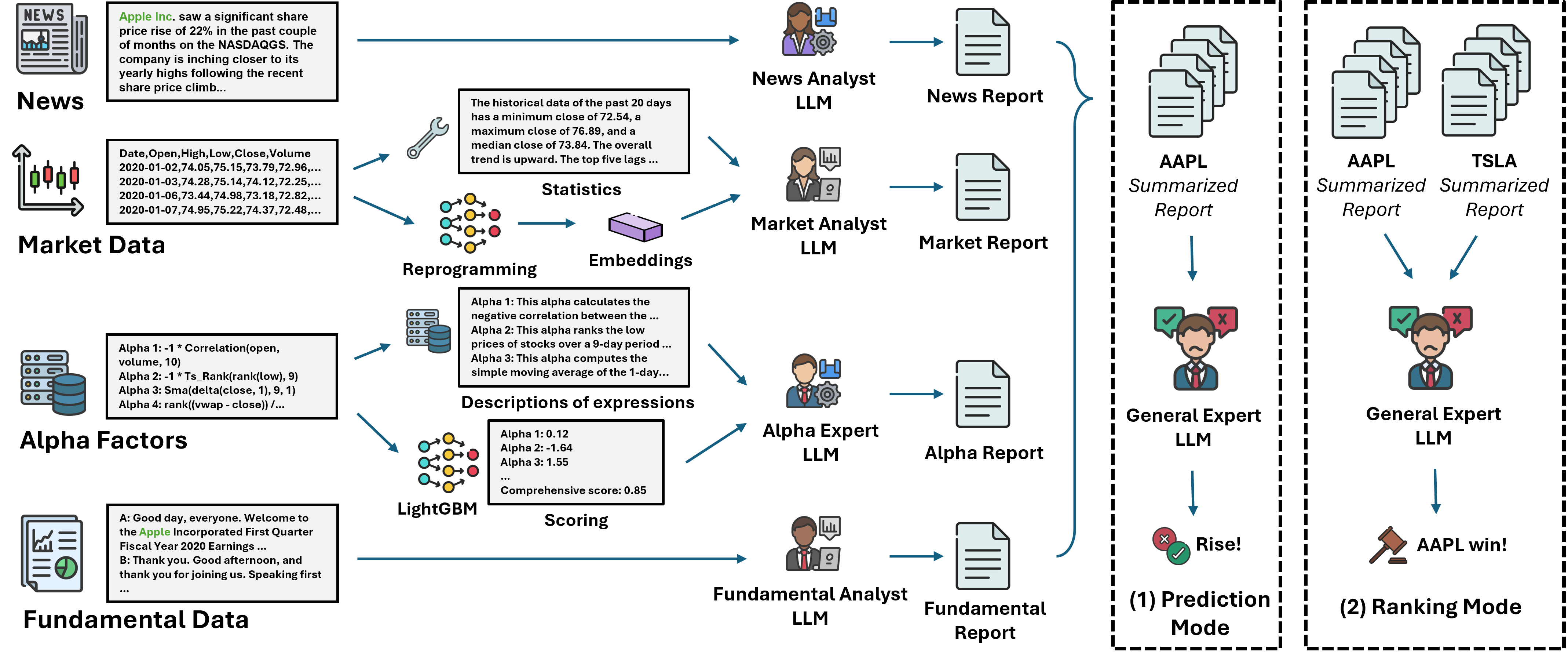}
\caption{TradExpert operates by processing distinct sources of financial data such as news texts, market data, alpha factors, and fundamental data through specialized expert LLMs. Then their reports are sumarized and sent to a General Expert which delivers the final outputs: (1) prediction of stock movement with prediction mode, (2) which of the two stocks is better or worse with ranking mode.  }
\label{fig:tradexpert_architecture}
\end{figure*}

\section{Problem Definition}

In this study, we aim to trade stocks using a framework that incorporates Large Language Models (LLMs).

The input data comprises four primary components:

\begin{itemize}
    \item \textbf{News:} Textual information from news articles pertinent to the stock and market conditions.
    \item \textbf{Market Data:} Historical OHLCV (Open, High, Low, Close, Volume) data representing the stock's trading activity.
    \item \textbf{Alpha Factors:} Quantitative indicators and signals believed to possess predictive power regarding stock price movements.
    \item \textbf{Fundamental Data:} Earnings call transcripts and fundamental metrics reflecting the company's economic health and performance.
\end{itemize}

\paragraph{Task 1: Stock movement prediction} is a fundamental challenge in quantitative trading, which involves the prediction of future price trends based on multifaceted data sources. Formally, let \( \mathcal{D} = \{ (x_i, y_i) \}_{i=1}^N \) denote our dataset, where \( x_i \) represents the input vector for the \( i \)-th stock on day \( t \), and \( y_i \in \{ \text{Rise}, \text{Fall} \} \) is the corresponding label indicating whether the stock price will rise or fall on day \( t+1 \). The input \( x_i \) can be expressed as:
\begin{align}
    x_i = \{ \text{News}_i, \text{Market}_i, \text{Factors}_i, \text{Fundamental}_i \}
    \label{eq:x}
\end{align}

Our objective is to learn a predictive function \( f \) parameterized by \( \theta \) such that $f_\theta(x_i) \approx y_i$, where \( f_\theta \) is modeled using LLMs. The model outputs a binary prediction, ``Rise'' or ``Fall'' indicating the predicted stock price movement.

\paragraph{Task 2: Stock trading simulation} involves evaluating the performance of \emph{Buy-and-Hold} strategy based on the Top-K ranked stocks sorted by TradExpert. This task simulates real-market trading scenarios to assess the profitability and risk of TradExpert using metrics including Annualized Return (AR), Sharpe Ratio (SR), Annualized Volatility (AV), and Maximum Drawdown (MD).

\section{Datasets}

In this study, we collected a comprehensive datasets encompassing various data sources including four primary components: News, Market Data, Alpha Factors, and Fundamental Data. The period covered by all data sources spans 4 years from January 1, 2020, to December 31, 2023. 

\subsection{Stastics}

\begin{wraptable}{r}{0.5\textwidth}
\centering
\small
\caption{Components in each data source. $\dagger$ and $\ddagger$ denote generated by GPT-4 and external models, respectively. }
\label{tab:pre_data}
\begin{tabular}{@{}lll@{}}
\toprule
\textbf{}            & \textbf{Prompts}                                                                                         & \textbf{Responses}                                                \\ \midrule
\textbf{News}        & News articles                                                                                                             & \begin{tabular}[c]{@{}l@{}}Movement\\ Reasoning$^\dagger$ \end{tabular} \\ \midrule
\textbf{Market}      & \begin{tabular}[c]{@{}l@{}}OHLCV embeddings\\ Statistics\end{tabular}                                               & Movement                                                          \\ \midrule
\textbf{Alpha}       & \begin{tabular}[c]{@{}l@{}}Expressions \\ Descriptions$^\dagger$ \\ Comprehensive score $^\ddagger$ \end{tabular} & Movement                                                          \\ \midrule
\textbf{Fundamental} & \begin{tabular}[c]{@{}l@{}}Earning calls scripts\\ Fundamental metrics\end{tabular}                                       & \begin{tabular}[c]{@{}l@{}}Movement\\ Reasoning$^\dagger$ \end{tabular} \\ \bottomrule
\end{tabular}
\end{wraptable}

\paragraph{News}

is collected from several reputable financial news sources, including Yahoo Finance, Reuters, InvestorPlace, GlobeNewswire, The Motley Fool, etc. This dataset comprises a total of 524,995 news articles for stocks on S\&P 500 list, with an average word count of 596.4 words per article. Each news article is associated with a list of related stock tickers. 

\paragraph{Market Data}

consists of historical daily OHLCV records for stocks on S\&P 500 list. This dataset includes a total of 481,484 records, offering a detailed view of the stocks' trading activity over the specified period.

\paragraph{Alpha Factors}

incorporates 108 technical indicators and factors with their expressions, which are believed to possess predictive power regarding stock price movements. 

\paragraph{Fundamental Data}

includes earnings call transcripts, financial statements, and fundamental metrics. The earnings call transcripts are sourced from Seeking Alpha, with 16 transcripts (4 years, quaterly updated) available for each stock. Fundamental metrics include Earnings Per Share (EPS), Price-to-Earnings Ratio (P/E Ratio), Book Value Per Share (BVPS), etc.

\subsection{Data Split}

The datasets were split into training, validation, and test sets based on chronological order to ensure that future data remains unseen during the training process. The split was performed as follows: Training set: January 1, 2020, to June 30, 2022. Validation set: July 1, 2022, to December 31, 2022. Testing set: January 1, 2023, to December 31, 2023.

\section{Methodology}

In this study, we propose \textbf{TradExpert}, a novel framework leveraging the MoE LLMs approach, where four LLMs serve as specialized experts for distinct sources of financial data. A General Expert LLM then synthesizes the summaries of the four Expert LLMs to produce the final output. The pipeline of TradExpert is shown in Figure~\ref{fig:tradexpert_architecture}.

In TradExpert, all expert LLMs are built on the LLaMA-2-7B backbone LLM~\cite{touvron2023llama} and are supervised and fine-tuned using the LoRA mechanism~\cite{hu2022lora}. Before training and fine-tuning, we preprocess the raw datasets to construct prompts, instructions, and ground-truth responses for each LLM. An overall description of the preprocessed datasets is demonstrated in Table~\ref{tab:pre_data}. The details will be introduced in the following.

\subsection{News Analyst}

The News Analyst LLM is designed to analyze texts of news articles to predict stock movements. The prompt and instruction for fine-tuning the LLM are shown in Figure~\ref{fig:news_analyst_combined_prompt}. The outputs from the News Analyst LLM include not only a prediction of the stock movement but also a reasoning of how the news article relates to the predicted movement in order to employ a Chain-of-Thought (CoT)~\cite{wei2022chain} reasoning approach. The ground-truth reasonings are pre-generated by the OpenAI GPT-4 API using instructions and prompts that incorporate the actual stock movements and the texts of news articles.

\begin{figure}[t]
\centering
\begin{tcolorbox}[colback=white, colframe=black, width=\columnwidth, arc=0mm, outer arc=0mm, boxrule=0.5mm, fontupper=\footnotesize, left=1mm, right=1mm, top=1mm, bottom=1mm]
\textbf{Instruction:} You are provided with a news article. Please predict how the stock will perform in the next \texttt{<D>} days. Your response should include your reasoning followed by a prediction of "Rise" or "Fall" in the specified format.

Format your response as follows:
Reasoning: [Your reasoning here]
Prediction: [Rise or Fall]

---

\textbf{Prompt:} News Article:
[Insert news article text here]

Question: Given the information in the news article above, how is the stock expected to perform in the next \texttt{<D>} days?

\end{tcolorbox}
\caption{Instruction and prompt for the News Analyst.}
\label{fig:news_analyst_combined_prompt}
\end{figure}

\subsection{Market Analyst}

The Market Analyst LLM focuses on analyzing historical OHLCV (Open, High, Low, Close, Volume) data to predict stock movements. However, time series data is inherently continuous and lacks the discrete token structure that LLMs are designed to process. This misalignment poses a significant challenge in effectively utilizing LLMs on time series. To this end, we utilize a reprogramming mechanism~\cite{jintime} to reprogram the input financial time series into text prototype representations.

Formally, let an OHLCV data instance be \( \mathbf{X}^{(i)} \in \mathbb{R}^{N \times T} \) which consists of \( N \) variables across \( T \) time steps. \( \mathbf{X}^{(i)} \) is first divided and embedded into a sequence of patch embeddings \( \left\{ \mathbf{X}_P^{(i)} \in \mathbb{R}^{N \times L_P \times d_m} \right\} \), where \( L_P\) and \( d_m \) are the number of patches and the patch embedding dimension respectively. The patches are then reprogrammed using a collection of text prototypes \( \mathbf{E}' \in \mathbb{R}^{V' \times D} \), which is achieved by linearly probing the LLM's pre-trained word embedding \( \mathbf{E} \in \mathbb{R}^{V \times D} \), where \( V \) and \(V'\) are the size of the vocabulary of the LLM and the text prototypes ( \( V' \ll V \)), and \( D \) is the embedding dimension. The reprogrammed patches are generated using a multi-head cross-attention mechanism: \( \mathbf{Z}_k^{(i)} = \text{Softmax}\left(\frac{\mathbf{Q}_k^{(i)} \mathbf{K}_k^{(i) \top}}{\sqrt{d_k}}\right) \mathbf{V}_k^{(i)}\), where query \( \mathbf{Q}_k^{(i)} =  \mathbf{X}_P^{(i)} \mathbf{W}_k^Q\), key \( \mathbf{K}_k^{(i)} = \mathbf{E}' \mathbf{W}_k^K\), and value \( \mathbf{V}_k^{(i)} = \mathbf{E}' \mathbf{W}_k^V\) for each head \( k \). The reprogrammed embeddings \( \mathbf{O}^{(i)} \) are obtained by aggregating the outputs from each attention head and projecting them to the hidden dimensions of the backbone LLM. 
Finally, the reprogrammed embeddings are augmented with a language description of statistics extracted from TSFresh~\cite{christ2018time}, serving as prompts for the Market Analyst. An example of instruction and prompt is shown in Figure~\ref{fig:market_analyst_fine_tuning_prompt}.

\begin{figure}[t]
\centering
\begin{tcolorbox}[colback=white, colframe=black, width=\columnwidth, arc=0mm, outer arc=0mm, boxrule=0.5mm, fontupper=\footnotesize, left=1mm, right=1mm, top=1mm, bottom=1mm]

\textbf{Instruction:} You are provided with historical OHLCV data of the past 20 days and a description of its statistics. Please predict how the stock will perform the next \texttt{<D>} day. Your response should be ``Rise" or "Fall".

---

\textbf{Prompt: }
\texttt{<Embbedings of reprogrammed OHLCV>}

Statistics: The historical prices have a minimum close of \texttt{<min\_val>} \texttt{<min\_D>} days ago, a maximum close of \texttt{<max\_val>} \texttt{<max\_D>} days ago, and a median close of \texttt{<median\_val>} \texttt{<median\_D>} days ago. The overall trend is \texttt{<upward or downward>}... 

Question: Given the reprogrammed OHLCV data and its statistics, how is the stock expected to perform in the next \texttt{<D>} days?
\end{tcolorbox}
\caption{Instruction and prompt for the Market Analyst.}
\label{fig:market_analyst_fine_tuning_prompt}
\end{figure}

\subsection{Alpha Expert}

The Alpha Expert specializes in processing expression-based alpha factors, which are technical indicators and algorithm-generated factors believed to possess predictie power regarding stock price movements.

We leverage GPT-4's capability of understanding complex expressions to pre-generate a language description for each factor. In this way, we built our Alpha database, where an alpha record consists of:
\begin{itemize}
    \item \textbf{Expression}: The mathematical or logical formula used to compute the alpha factor based on OHLCV data. E.g.  \texttt{rank(ts\_argmax(corr(ts\_rank(close, 10), ts\_rank(volume, 10), 10), 5))}
    \item \textbf{Description}: Generated by GPT-4 with prompts that include the expression.
\end{itemize}


For each stock, we first calculate the values of all alpha factors based on OHLCV data and then derive a comprehensive score via a LightGBM-based model~\cite{ke2017lightgbm}. Subsequently, we select Top-K alphas that contribute most significantly to this comprehensive score. Descriptions of these Top-K alphas are retrieved from the database and, along with the calculated values, are included in the prompts and instructions for the Alpha Expert.

\subsection{Fundamental Analyst}

The Fundamental Analyst LLM specializes in analyzing fundamental data, including earnings call transcripts and financial metrics, to predict stock price movements on a quarterly basis. The procedure of the Fundamental Analyst LLM is similar to that of the News Analyst LLM, with key differences being that the fundamental data is updated quarterly and, therefore, the movement predictions are made for the next quarter. The response should include a prediction in one of the following five categories: ``Strong Rise", ``Moderate Rise", ``No Change", ``Moderate Fall", or ``Strong Fall", followed by a reasoning.

\subsection{General Expert}

The General Expert LLM can operate in two distinct modes: prediction mode and ranking mode. Both modes begin by summarizing the reports (historical conversation including instructions, prompts, and responses) from the four specialized experts due to the limitations on input context length of the backbone LLM.

In \textbf{prediction mode}, used for stock movement prediction, the summarized reports are used to construct a prompt with a prediction prefix. Given the summarized reports, the General Expert LLM outputs a binary prediction indicating whether the stock will rise or fall. 

In \textbf{ranking mode}, used for stock trading, the General Expert LLM functions as a comparator to establish the ranking ability. Specifically, given the summarized reports of two stocks, the General Expert LLM would determine which stock is likely to perform better in the future. To generate a Top-K ranking of stocks, we employ a relaxed comparison-based sorting similar to BubbleSort: We initially compare every pair of stocks and count the number of wins for each stock. Subsequently, we sort these counts to establish the rankings for stocks. Although algorithms like QuickSort and vanilla BubbleSort offer fewer comparisons for Top-K selection on average $\mathcal{O}(N \log N)$ and $\mathcal{O}(N\cdot K)$, we propose to use this relaxed comparison-based sorting alogrithm with $\mathcal{O}(N^2)$ due to the non-transitive nature of LLM-based comparator~\cite{liu2024aligning}. Therefore, more comparisons tend to yield more accurate rankings in practice.

The General Expert LLM is finetuned on both tasks of stock movement prediction and stock comparison simultaneously. The instructions and prompts are shown in Figure~\ref{fig:general_expert_prompts}.

\begin{figure}[t]
\centering
\begin{tcolorbox}[colback=white, colframe=black, width=\columnwidth, arc=0mm, outer arc=0mm, boxrule=0.5mm, fontupper=\footnotesize, left=1mm, right=1mm, top=1mm, bottom=1mm]

\textbf{Instruction:} You are provided with a summarized report of the stock. Please predict whether the stock will rise or fall the next \texttt{<D>} day.

Format your response as follows: Reasoning: [Your reasoning here] Prediction: [Rise or Fall].

---

\textbf{Prompt:} Summarized Report:
[Insert summarized report here]

Question: Based on the summarized report, will the stock rise or fall in the next $<D>$ days?

\vspace{0.2cm}
\hrule
\vspace{0.2cm}

\textbf{Instruction:} You are provided with summarized reports of two stocks. Please determine which stock will perform better the next \texttt{<D>} day. Please output Stock \texttt{AAA} or Stock \texttt{BBB}.

---

\textbf{Prompt:} Summarized Report for Stock \texttt{AAA}:
[Report A]

Summarized Report for Stock \texttt{BBB}:
[Report B]

Question: Based on the summarized reports, which stock will perform better in the next $<D>$ days?
\end{tcolorbox}
\caption{Instructions and prompts for the General Expert LLM: (Top) Prediction mode, (Bottom) Ranking mode.}
\label{fig:general_expert_prompts}
\end{figure}

\begin{table*}[t]
\small
\centering
\caption{Comparison results on stock movement prediction task. As a binary classification problem, methods are evaluated by Accuracy (Acc) and Mattheus Correlation Coefficient (MCC). Both metrics are better with higher values. The best and second best results are in \textbf{bold} and \underline{underlined}, respectively.}
\label{tab:my-table}
\begin{tabular}{@{}lcccccccc@{}}
\toprule
\textbf{}                 & \multicolumn{2}{c}{\textbf{BigData22}} & \multicolumn{2}{c}{\textbf{ACL18}} & \multicolumn{2}{c}{\textbf{CIKM18}} & \multicolumn{2}{c}{\textbf{S\&P500 (Ours)}} \\ \midrule
\textbf{}                 & \textbf{Acc ↑}       & \textbf{MCC ↑}      & \textbf{Acc ↑}     & \textbf{MCC ↑}    & \textbf{Acc ↑}     & \textbf{MCC ↑}     & \textbf{Acc ↑}         & \textbf{MCC ↑}         \\ \midrule
\textbf{}                 & \multicolumn{8}{l}{\textbf{Hybrid Models}}                                                                                                                      \\
\textbf{ALSTM-W}          & 0.48               & -0.01             & 0.53             & 0.08            & 0.54             & 0.03             & 0.55                 & 0.10                 \\
\textbf{ALSTM-D}          & 0.49               & 0.01              & 0.53             & 0.07            & 0.50             & -0.04            & 0.54                 & 0.05                 \\
\textbf{StockNet}         & 0.53               & -0.02             & 0.54             & -0.02           & 0.52             & -0.02            & 0.56                 & 0.06                 \\
\textbf{SLOT}             & 0.55               & 0.10              & \underline{0.59}             & \textbf{0.21}            & 0.56             & \underline{0.09}             & /                    & /                    \\ \midrule
\textbf{}                 & \multicolumn{8}{l}{\textbf{Large Language Models}}                                                                                                              \\
\textbf{GPT-4}            & 0.54               & 0.03              & 0.52             & 0.02            & 0.57             & 0.02             & 0.58                 & 0.10                 \\
\textbf{Gemini}           & 0.55               & 0.04              & 0.52             & 0.04            & 0.54             & 0.02             & 0.59                 & \underline{0.11}                 \\
\textbf{LLaMA2-7B-chat}   & 0.54               & 0.05              & 0.51             & 0.01            & 0.55             & -0.03            & 0.57                 & 0.06                 \\
\textbf{LLaMA2-70B}       & 0.47               & 0.00              & 0.51             & 0.01            & 0.49             & -0.07            & 0.57                 & 0.02                 \\
\textbf{LLaMA3-8B}        & 0.55               & 0.02              & 0.52             & 0.02            & \underline{0.57}             & 0.03             & 0.53                 & 0.04                 \\
\textbf{FinMA-7B}         & 0.51               & 0.02              & 0.51             & 0.03            & 0.50             & 0.08             & 0.59                 & 0.09                 \\
\textbf{FinGPT-7B-lora}   & 0.45               & 0.00              & 0.49             & 0.00            & 0.42             & 0.00             & 0.51                 & 0.05                 \\
\textbf{InternLM}         & \underline{0.56}               & \underline{0.08}              & 0.51             & 0.02            & \underline{0.57}             & -0.03            & \underline{0.60}                 & 0.06                 \\
\textbf{Falcon}           & 0.55               & 0.00              & 0.51             & 0.00            & 0.47             & -0.06            & 0.52                 & 0.03                 \\
\textbf{Mixtral-7B}       & 0.46               & 0.02              & 0.49             & 0.00            & 0.42             & -0.05            & 0.54                 & 0.04                 \\ \midrule
\textbf{}                 & \multicolumn{8}{l}{\textbf{MoE Larage Language Models}}                                                                                                         \\
\textbf{TradExpert-NM} & \textbf{0.59}               & \textbf{0.12}              & \textbf{0.60}             & \underline{0.15}            & \textbf{0.59}             & \textbf{0.12}             & \textbf{0.64}                 & \textbf{0.19}                 \\ \bottomrule
\end{tabular}
\end{table*}

\section{Experiments}

In this section, we conduct a comprehensive evaluation for TradExpert framework on two main tasks: stock movement prediction and stock trading simulation. Our experiments aims to address the following research questions: \textbf{RQ1:} How does TradExpert perform in stock movement prediction compared with state-of-the-art baselines? \textbf{RQ2:} What are the potential profits and associated risks of TradExpert in the backtesting on the real market? \textbf{RQ3:} How effective is the reasoning capability of TradExpert for unstructured data? \textbf{RQ4:} What is the significance of each expert within the TradExpert framework? \textbf{RQ5:} Why we choose the relaxed comparison-based sorting algorithm in TradExpert?

\subsection{Datasets}

We include two categories of datasets in our experiments:

\begin{itemize}
    \item \textbf{Benchmark Datasets:} We use publicly available benchmark datasets in stock movement prediction research including CIKM18~\cite{wu2018hybrid}, ACL18~\cite{xu2018stock}, and BigData22~\cite{soun2022accurate} datasets.
    \item \textbf{Proprietary Datasets:} We also utilize our proprietary datasets, which include extensive historical OHLCV data, news articles, alpha factors, and fundamental metrics for a comprehensive analysis.
\end{itemize}

\subsection{Experimental Setup}

In our experiments, the four expert LLMs and the General Expert LLM are bulit on the LLaMA-2-7B bakcbone model~\cite{touvron2023llama} and are finetuned via LoRA~\cite{hu2022lora} mechanism.

\paragraph{Stock Movement Prediction:} TradExpert works in prediction mode, that is, the General Expert LLM reponses a binary prediction indicating whether a stock will rise or fall the next day. Methods are evaluated using binary classification metrics such as accuracy (Acc) and Matthews Correlation Coefficient (MCC).

\paragraph{Stock Trading Simulation:} TradExpert works in ranking mode, that is, the General Expert LLM acts as a comparator to sort the stocks. We simulate the real profit and risk of TradExpert by executing trades based on the Top-K ranked stocks. TradExpert and baselines are evaluated using metrics including Annualized Return (AR), Sharpe Ratio (SR), Annualized Volatility (AV), and Maximum Drawdown (MD). 

\subsection{Baselines}

For stock movement prediction, the baselines include: \textbf{(1) Hybrid Models:} StockNet~\cite{xu2018stock}, ALSTM-W~\cite{soun2022accurate}, ALSTM-D~\cite{soun2022accurate}, SLOT~\cite{soun2022accurate}. \textbf{2) Large Language Models:} GPT-4~\cite{achiam2023gpt}, Gemini~\cite{team2023gemini}, LLaMA2-70B~\cite{touvron2023llama}, LLaMA3-8B~\cite{llama3modelcard}, FinMA-7B~\cite{xie2023pixiu}, FinGPT-LlaMA2-7B~\cite{yang2023fingpt}, InternLM-7B~\cite{cai2024internlm2}, Falcon-7B~\cite{almazrouei2023falcon}, Mixtral-7B~\cite{jiang2023mistral}.

For stock trading simulation, the baselines include: \textbf{(1)Traditional Models:} Random Forest~\cite{breiman2001random}, Decision Tree~\cite{loh2011classification}, SVM~\cite{cortes1995support}. \textbf{(2) Deep Learning Models:}  A2C~\cite{mnih2016asynchronous}, PPO~\cite{schulman2017proximal}, SARL~\cite{ye2020reinforcement}, EIIE~\cite{jiang2017deep}, and DeepTrader~\cite{wang2021deeptrader}. To reduce computational costs in backtesting, we evaluated all methods on datasets with stocks on the DOW 30 list, a subset of the S\&P 500, with around 30 stocks.

\begin{table}
\small
\centering
\caption{Comparison results on stock trading simulation task with stocks on the DOW 30. Annualized Return (AR), Sharpe Ratio (SR), Annualized Volatility (AV), and Maximum Drawdown (MD) are utilized to evaluate the profits and risks of methods. The best results are in \textbf{bold}.}
\label{tab:trading}
\begin{tabular}{@{}lllll@{}}
\toprule
\textbf{}           & \textbf{AR ↑} & \textbf{AV ↓} & \textbf{SR  ↑} & \textbf{MD ↓} \\ \midrule
\textbf{DJI Index}  & 13.92\%                           & 11.41\%                               & 1.22                          & 9.70\%                       \\ \midrule
\textbf{}           & \multicolumn{4}{l}{\textbf{Traditional Models}}                                                                                          \\
\textbf{SVM}        & 15.77\%                           & 26.67\%                               & 0.59                          & 19.94\%                      \\
\textbf{XGBoost}    & 21.58\%                           & 27.29\%                               & 0.79                          & 21.90\%                      \\
\textbf{LightGBM}   & 2.17\%                            & 22.74\%                               & 0.1                           & 21.29\%                      \\ \midrule
\textbf{}           & \multicolumn{4}{l}{\textbf{Deep Learning Models}}                                                                                        \\
\textbf{A2C}        & 19.16\%                           & 11.29\%                               & 1.7                           & 9.09\%                       \\
\textbf{PPO}        & 16.62\%                           & 11.51\%                               & 1.44                          & 9.45\%                       \\
\textbf{EIIE}       & 23.64\%                           & 13.73\%                               & 1.72                          & 10.07\%                      \\
\textbf{SARL}       & 21.87\%                           & 14.72\%                               & 1.49                          & 8.52\%                       \\
\textbf{DeepTrader} & 32.45\%                           & 17.86\%                               & 1.82                          & 15.32\%                      \\ \midrule
\textbf{}           & \multicolumn{4}{l}{\textbf{MoE Large Language Models}}                                                                                   \\
\textbf{TradExpert} & \textbf{49.79\%}                           & \textbf{9.95\%}                                & \textbf{5.01}                          & \textbf{6.56\%}                       \\ \bottomrule
\end{tabular}
\end{table}

\subsection{Results}

\subsubsection{Stock Movement Prediction}

We implemented all baselines ourselves or utilized existing open-source codes, except the closed-source model SLOT, for which we refer to the metrics reported in the relevant paper. To ensure a fair comparison, we only included the News Analyst and Market Analyst in TradExpert, named TradExeprt-NM. The results are shown in Table~\ref{tab:my-table}. As we can see, among hybrid models, SLOT achieves outstanding accuracy and MCC on the ACL18, benefitting from the proposed global market guidance. Among LLMs, InternLM shows remarkable performance, particularly on our proprietary S\&P500 dataset. Our proposed TradExpert-NM, utilizing a mixture of expert LLMs approach, consistently outperformed other models across all datasets except for MCC on the ACL18, showcasing its superior performance. Noting that BigData22, ACL18, and CIKM18 are relatively small datasets with texts from tweets, while our S\&P500 dataset consist of news articles with much more words. This difference in text lengths contributes to the more significant improvements obtained by TradExpert-7B-NM on the S\&P500 dataset.

\subsubsection{Stock Trading Simulation}

\begin{wrapfigure}{r}{0.55\textwidth}
  \centering
  \includegraphics[width=1.0\linewidth]{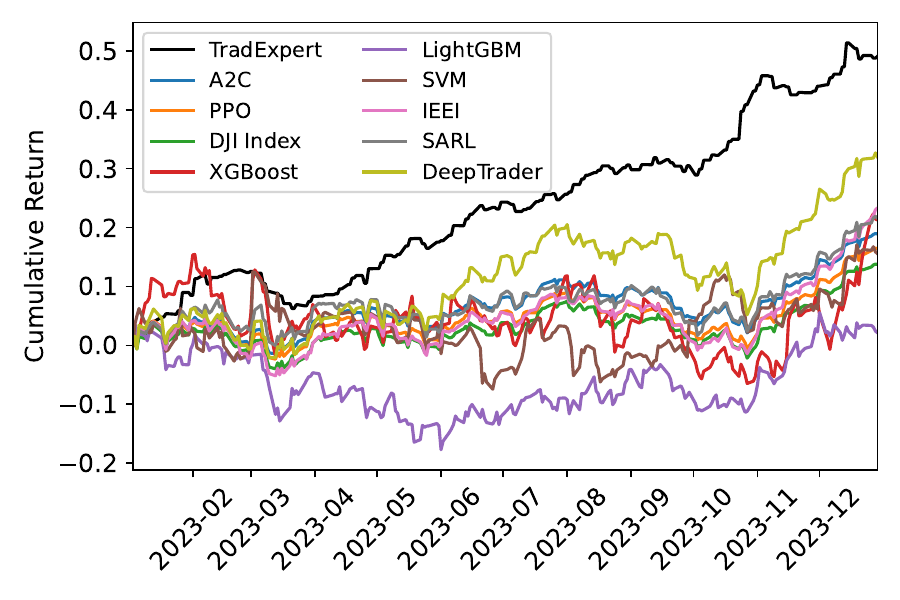}
  \caption{Cumulative returns over time of all methods on 30 stocks on DOW 30 list. DJI Index represents the market trend.}
  \label{fig:all_bt}
\end{wrapfigure}

We perform backtesting to evaluate TradExpert and baselines. To reduce computational costs in backtesting, we limit the stock pool to about 30 stocks on the DOW 30, a subset of the S\&P 500. For TradExpert, we implement a \emph{Buy-and-Hold} trading strategy on the Top-K stocks ranked by TradExpert. The backtesting period is the same as the testing period of our datasets, which ranges from January 1, 2023, to December 31, 2023. The results summarized in Table~\ref{tab:trading} demonstrate TradExpert's superior performance across all metrics considered. Among traditional models, XGBoost achieved a relatively high return but also exhibited high volatility and drawdown, indicating greater risk. Deep learning models generally outperformed traditional models. Among them, DeepTrader stood out with the highest return and Sharpe ratio. TradExpert, our proposed model, significantly outperformed all other models with an exceptional AR of 49.79\% and the lowest AV of 9.95\%. This combination yielded an outstanding Sharpe ratio of 5.01, indicating a high return per unit of risk. Figure~\ref{fig:all_bt} shows the trends of cumulative returns over time for all methods.

\section{Ablation Study}


\paragraph{The Impacts of Experts}

\begin{wraptable}{r}{0.55\textwidth}
\small
\centering
\caption{Ablation study for the impacts of experts.}
\label{tab:ab_expert}
\begin{tabular}{@{}lllll@{}}
\toprule
\textbf{Configuration}      & \textbf{AR ↑} & \textbf{AV ↓} & \textbf{SR ↑} & \textbf{MD ↓} \\ \midrule
\textbf{TradExpert}         & 49.79\%       & 9.95\%        & 5.01          & 6.56\%        \\ \midrule
\textbf{w/o Market}         & 30.87\%       & 16.43\%       & 1.88          & 13.29\%       \\
\textbf{w/o News}           & 31.92\%       & 18.36\%       & 1.74          & 13.04\%       \\
\textbf{w/o Alpha}          & 41.65\%       & 11.38\%       & 3.66          & 8.94\%        \\
\textbf{w/o Fundamental}    & 44.32\%       & 10.68\%       & 4.15          & 7.82\%        \\ \bottomrule
\end{tabular}
\end{wraptable}

To evaluate the effectiveness of each expert within the TradExpert framework, we created multiple versions of TradExpert, each with a specific expert removed. By comparing the performance of these modified frameworks, we can assess the impact of each expert on the overall functionality of TradExpert. The results in Table~\ref{tab:ab_expert} reveal the varying degrees of impact of each expert. The Market Analyst and the News Analyst emerged as the most critical, significantly influencing profitability and risk management, as seen by the largest drop in AR and AV when they were removed, respectively. The Alpha Expert is obviously less impactful than the Market Analysts and the News Analysts. The Fundamental Analyst had the smallest effect on daily trading metrics, but provided essential long-term stability, evident from the modest changes in AR and MD upon its removal. This highlights a strategic balance in TradExpert, where each expert contributes uniquely to the final decision and prediction.

\paragraph{The Effectiveness of Structured Data Reasoning.} 

\begin{wraptable}{r}{0.5\textwidth}
\centering
\caption{Ablation study for the effectiveness of structured data reasoning in predicting day $T+1$'s returns.}
\label{tab:alpha}
\small
\begin{tabular}{lcccc}
\toprule
\textbf{Model}  & \textbf{RankIC ↑} & \textbf{RankICIR ↑} \\
\midrule
\textbf{TradExpert-MA}  & 0.12 & 0.90 \\
\textbf{Alpha Combination}  & 0.07 & 0.65 \\
\bottomrule
\end{tabular}
\end{wraptable}

We show the effectiveness by comparing TradExpert-MA with traditional models for structured data like OHLCV data and alpha factors. We use a genetic programming-based symbolic regression model as our baseline, which mines alpha expressions aimed at predicting the RankIC of day $T+1$'s returns. TradExpert-MA is built on top of the same alphas, where News and Fundamental experts were removed to exclude affects from other sources. We compare TradExpert-MA with the combination of alphas using metrics of RankIC and RankICIR. The results are shown in Table~\ref{tab:alpha}. The improvements over the alpha combination demonstrate the reasoning ability of TradExpert for structured data.

\paragraph{The Choices of Ranking Algorithm}

\begin{wraptable}{r}{0.6\textwidth}
\small
\centering
\caption{Ablation study for the choices of ranking algorithm. $\dagger$ denotes being equipped in TradExpert.}
\label{tab:ab_rank}
\begin{tabular}{lccc}
\toprule
 & \textbf{RankIC ↑} & \textbf{RankICIR ↑} & \textbf{Time} \\
\midrule
\textbf{RelaxedSort$^\dagger$}      & 0.12 & 0.90 & \(O(N^2)\) \\ \midrule
\textbf{BubbleSort}    & 0.06 & 0.65 & \(O(N\cdot K)\) \\
\textbf{QuickSort}          & 0.03 & 0.38 &  \(O(N \log N)\) \\
\bottomrule
\end{tabular}
\end{wraptable}

In TradExpert, we implement the Top-K ranking by sorting all stocks completely using a relaxed comparison-based algorithm, where TradExpert serves as the comparator. To justify our choice of this seemingly cumbersome approach, we conducted comparison experiments with other theoretically more efficient ranking algorithms. Specifically, our alternatives include QuickSort and BubbleSort with time complexity $\mathcal{O}(N \log N)$ and $\mathcal{O}(N\cdot K)$, respectively. The comparison results in Table~\ref{tab:ab_rank} demonstrate that our approach outperforms others, despite having a higher computational complexity. This is attributed to the non-transitive nature of LLM-based comparator. Therefore, a greater number of comparisons yield more accurate rankings in TradExpert.

\section{Conclusion}

In this study, we introduced TradExpert, a novel framework that harnesses the power of LLMs to enhance stock trading strategies. By integrating multiple specialized LLMs, each focused on distinct aspects of financial data, TradExpert provides a comprehensive and nuanced analysis that significantly outperforms traditional financial models in practice. Looking ahead, our goal is to explore how to employ TradExpert in the high-frequency trading scenario and extend its capabilities to encompass a wider range of global markets. 

\paragraph{Limitation} Although TradExpert has notable strengths, its processing time poses certain challenges. On average, it takes 4.7 seconds for a single stock with an Nvidia A5000 GPU. For daily trading, this processing time is generally manageable. However, for scenarios demanding quicker decision-making, such as high-frequency trading, TradExpert's latency becomes a notable drawback.

\bibliography{ref}
\bibliographystyle{iclr2025_conference}

\appendix
\section{Implementation Details}

\subsection{Environment}
All experiments were conducted on a machine with the following specifications:
\begin{itemize}
    \item \textbf{Hardware:} NVIDIA A5000 GPU x 4, AMD Ryzen Threadripper PRO 3975WX CPU, 256 GB RAM
    \item \textbf{Software:} Ubuntu 22.04, Python 3.9, PyTorch 2.2, CUDA 12.3
\end{itemize}

\subsection{Training}
The proposed TradExpert framework utilizes a Mixture of Experts (MoE) approach, where four LLMs are specialized in processing distinct sources of financial data. All these LLMs are based on the LLaMA-2-7B~\cite{touvron2023llama} model and fine-tuned using the LoRA mechanism~\cite{hu2022lora}. Below are the specifics:

\begin{itemize}
    \item \textbf{Expert LLMs:} Each expert LLM specializes in one of the following data sources: News, Market Data, Alpha Factors, and Fundamental Data.
    \item \textbf{General Expert LLM:} Integrates outputs from the four expert LLMs to make final predictions or rankings.
    \item \textbf{Fine-tuning:} We employ supervised-fine-tuning (SFT) with LoRA for all LLMs.
\end{itemize}

\subsection{Hyperparameters}

The key hyperparameters used for each of the LLMs in the TradExpert framework, including those fine-tuned with LoRA, are summarized in Table~\ref{tab:hyperparameters}.

\begin{table*}[t]
\tiny
\centering
\caption{Hyperparameters for Expert LLMs and General Expert LLM in the TradExpert framework.}
\label{tab:hyperparameters}
\begin{tabular}{|l|c|c|c|c|c|}
\hline
\textbf{Hyperparameter}              & \textbf{News Analyst} & \textbf{Market Analyst} & \textbf{Alpha Expert} & \textbf{Fundamental Analyst} & \textbf{General Expert} \\ \hline
\textbf{Model Architecture}          & LLaMA-2-7B            & LLaMA-2-7B              & LLaMA-2-7B            & LLaMA-2-7B                  & LLaMA-2-7B             \\ \hline
\textbf{Learning Rate}               & $1e^{-4}$             & $1e^{-4}$               & $1e^{-4}$             & $1e^{-4}$                   & $1e^{-4}$              \\ \hline
\textbf{Batch Size}                  & 4                     & 4                       & 4                     & 4                           & 4                      \\ \hline
\textbf{Number of Epochs}            & 30                    & 30                      & 30                    & 30                          & 30                     \\ \hline
\textbf{Optimizer}                   & AdamW                 & AdamW                   & AdamW                 & AdamW                       & AdamW                  \\ \hline
\textbf{Sequence Length}             & 2048                  & 2048                    & 2048                  & 2048                        & 2048                   \\ \hline
\textbf{LoRA Rank}                   & 8                     & 8                       & 8                     & 8                           & 8                      \\ \hline
\textbf{LoRA Alpha}                  & 16                    & 16                      & 16                    & 16                          & 16                     \\ \hline
\textbf{LoRA Dropout}                & 0.1                   & 0.1                     & 0.1                   & 0.1                         & 0.1                    \\ \hline
\textbf{Warmup Ratio}                & 0.1                   & 0.1                     & 0.1                   & 0.1                         & 0.1                    \\ \hline
\textbf{Gradient Accumulation Steps} & 8                     & 8                       & 8                     & 8                           & 8                      \\ \hline
\end{tabular}
\end{table*}

\subsection{Evaluation Metrics}
The performance of the models was evaluated using the following metrics, with their respective equations and descriptions:

\subsubsection{Stock Movement Prediction}

\paragraph{Accuracy (Acc):}
Accuracy is the ratio of correctly predicted stock movements to the total number of predictions. It is defined as:
\begin{equation}
\text{Accuracy} = \frac{TP + TN}{TP + TN + FP + FN}
\end{equation}
where:
\begin{itemize}
    \item \( TP \) is the number of true positives (correctly predicted rises).
    \item \( TN \) is the number of true negatives (correctly predicted falls).
    \item \( FP \) is the number of false positives (incorrectly predicted rises).
    \item \( FN \) is the number of false negatives (incorrectly predicted falls).
\end{itemize}

\paragraph{Matthews Correlation Coefficient (MCC):}
The Matthews Correlation Coefficient is a balanced measure that considers all four quadrants of the confusion matrix (true/false positives and true/false negatives). It is especially useful for imbalanced datasets. MCC is defined as:
\begin{equation}
\text{MCC} = \frac{(TP \times TN) - (FP \times FN)}{\sqrt{(TP + FP)(TP + FN)(TN + FP)(TN + FN)}}
\end{equation}
MCC returns a value between -1 and 1, where 1 indicates perfect prediction, 0 indicates no better than random prediction, and -1 indicates total disagreement between prediction and observation.

\subsubsection{Stock Trading Simulation}

\paragraph{Annualized Return (AR):}
Annualized Return represents the geometric average amount of money earned by an investment each year over a given period. It is defined as:
\begin{equation}
\text{AR} = \left( \prod_{i=1}^{N} (1 + r_i) \right)^{\frac{252}{N}} - 1
\end{equation}
where \( r_i \) is the return on day \( i \) and \( N \) is the number of trading days. The factor of 252 is used to annualize the return (assuming there are 252 trading days in a year).

\paragraph{Sharpe Ratio (SR):}
The Sharpe Ratio measures the performance of an investment compared to a risk-free asset, after adjusting for its risk. It is defined as:
\begin{equation}
\text{SR} = \frac{\text{AR} - R_f}{\sigma}
\end{equation}
where:
\begin{itemize}
    \item \( \text{AR} \) is the annualized return of the portfolio.
    \item \( R_f \) is the risk-free rate (which is $0$ in this paper for simplicity).
    \item \( \sigma \) is the annualized standard deviation of daily returns (a measure of volatility).
\end{itemize}
A higher Sharpe Ratio indicates better risk-adjusted returns.

\paragraph{Annualized Volatility (AV):}
Annualized Volatility is a measure of the risk or uncertainty of a financial instrument. It quantifies the degree of variation of trading prices over time. It is defined as:
\begin{equation}
\text{AV} = \sigma \sqrt{252}
\end{equation}
where \( \sigma \) is the standard deviation of daily returns, and 252 is the number of trading days in a year. 

\paragraph{Maximum Drawdown (MD):}
Maximum Drawdown is the maximum observed loss from a peak to a trough of a portfolio, before a new peak is attained. It is defined as:
\begin{equation}
\text{MD} = \max_{t \in [0, T]} \left( \frac{\text{Peak}_t - \text{Trough}_t}{\text{Peak}_t} \right)
\end{equation}
where \( \text{Peak}_t \) is the maximum value at time \( t \) and \( \text{Trough}_t \) is the minimum value following the peak, within the observed time period \( [0, T] \). Maximum Drawdown gives insight into the risk of a portfolio by identifying the largest possible loss.

\subsection{Reproducibility}
To ensure the reproducibility of our results, we have included a partial of the pseudo codes and datasets used in our experiments in the supplementary material due to the limitation of file size.

\section{Algorithm}

We illustrate the algorithm pipeline of TradExpert framework and relaxed comparison-based sorting algorithm for ranking mode of General Expert.

\begin{algorithm}[H]
\caption{TradExpert Framework}
\label{alg:TradExpert}
\begin{algorithmic}[1]
\STATE \textbf{Input:} Financial datasets \\ $\mathcal{D} = \{News, Market, Alpha, Fundamental\}$
\STATE \textbf{Output:} Predicted stock movements or Top-K ranked stocks for trading

\STATE \textbf{Step 1: Data Preprocessing}
\STATE Preprocess each dataset:
\begin{itemize}
    \item Tokenize and truncate news articles to a fixed length.
    \item Reprogram market data (OHLCV) into text prototypes.
    \item Generate language descriptions for alpha factors.
    \item Tokenize earnings call transcripts and generate fundamental metrics prompts.
\end{itemize}

\STATE \textbf{Step 2: Expert LLMs Processing}
\FOR{each dataset in $\mathcal{D}$}
    \STATE Apply the corresponding Expert LLM:
    \begin{itemize}
        \item \textbf{News Analyst:} Analyze news articles and predict stock movement (CoT with reasoning).
        \item \textbf{Market Analyst:} Process market data and predict stock movement.
        \item \textbf{Alpha Expert:} Evaluate alpha factors and predict stock movement.
        \item \textbf{Fundamental Analyst:} Analyze fundamental data for quarterly predictions (CoT with reasoning).
    \end{itemize}
\ENDFOR

\STATE \textbf{Step 3: Integration by General Expert LLM}
\STATE Summarize the outputs of all Expert LLMs.
\STATE Use the General Expert LLM to:
\begin{itemize}
    \item \textbf{Prediction Mode:} Predict whether the stock will rise or fall.
    \item \textbf{Ranking Mode:} Compare and rank stocks to select the Top-K for trading.
\end{itemize}

\STATE \textbf{Step 4: Final Output}
\STATE Return the predicted stock movements or the Top-K ranked stocks for trading.

\end{algorithmic}
\end{algorithm}

\begin{algorithm}
\caption{Relaxed Comparison-Based Sorting Algorithm for Ranking}
\label{alg:relaxed_sorting}
\begin{algorithmic}[1]
\REQUIRE $S = \{s_1, s_2, \dots, s_n\}$ \COMMENT{Set of stocks}
\REQUIRE $R = \{r_1, r_2, \dots, r_n\}$ \COMMENT{Summarized reports for each stock}
\REQUIRE $K$ \COMMENT{Number of top stocks to select}

\STATE $C \gets [0, 0, \dots, 0]$ \COMMENT{Initialize win counts for each stock}

\FOR{$i = 1$ to $n$}
    \FOR{$j = i+1$ to $n$}
        \STATE $result \gets \text{General\_Expert\_LLM\_Compare}(r_i, r_j)$
        \IF{$result$ is True}
            \STATE $C[i] \gets C[i] + 1$ \COMMENT{Stock $s_i$ wins}
        \ELSE
            \STATE $C[j] \gets C[j] + 1$ \COMMENT{Stock $s_j$ wins}
        \ENDIF
    \ENDFOR
\ENDFOR

\STATE $ranked\_stocks \gets \text{Sort}(S, \text{key} = C, \text{reverse} = \text{True})$
\COMMENT{Sort stocks by win counts in descending order}

\STATE $top\_k\_stocks \gets \text{ranked\_stocks}[1:K]$ \COMMENT{Select the Top-$K$ stocks}

\RETURN $top\_k\_stocks$
\end{algorithmic}
\end{algorithm}

\section{Instructions and Prompts}

The TradExpert framework utilizes several specialized Large Language Models (LLMs) to process distinct types of financial data, each guided by specific instructions and prompts. 

\begin{itemize}
    \item \textbf{News Analyst LLM:} The instructions and prompt for predicting stock movements based on news articles are shown in Figure \ref{fig:news_analyst_combined_prompt}.
    \item \textbf{Market Analyst LLM:} The instructions and prompt for analyzing historical OHLCV data are shown in Figure \ref{fig:market_analyst_combined_prompt}.
    \item \textbf{Alpha Expert LLM:} The instructions and prompt for evaluating alpha factors derived from OHLCV data are shown in Figure \ref{fig:alpha_expert_combined_prompt}.
    \item \textbf{Fundamental Analyst LLM:} The instructions and prompt for predicting stock movements based on earnings call transcripts and fundamental metrics are shown in Figure \ref{fig:fundamental_analyst_combined_prompt}.
    \item \textbf{General Expert LLM (Prediction Mode):} The instructions and prompt for predicting stock movements by integrating outputs from other experts are shown in Figure \ref{fig:general_expert_prediction_mode}.
    \item \textbf{General Expert LLM (Ranking Mode):} The instructions and prompt for comparing and ranking stocks are shown in Figure \ref{fig:general_expert_ranking_mode}.
\end{itemize}

\begin{figure*}[t]
\centering
\begin{tcolorbox}[colback=white, colframe=black, width=1\columnwidth, arc=0mm, outer arc=0mm, boxrule=0.5mm, fontupper=\footnotesize, left=1mm, right=1mm, top=1mm, bottom=1mm]
\textbf{Instruction:} You are provided with a news article. Please predict how the stock will perform in the next \texttt{<D>} days. Your response should include your reasoning followed by a prediction of "Rise" or "Fall" in the specified format.

Format your response as follows:
Reasoning: [Your reasoning here]
Prediction: [Rise or Fall]

---

\textbf{Prompt:} News Article:
[Insert news article text here]

Question: Given the information in the news article above, how is the stock expected to perform in the next \texttt{<D>} days?

\end{tcolorbox}
\caption{Instruction and prompt for the News Analyst.}
\label{fig:news_analyst_combined_prompt}
\end{figure*}

\begin{figure}[t]
\centering
\begin{tcolorbox}[colback=white, colframe=black, width=1\columnwidth, arc=0mm, outer arc=0mm, boxrule=0.5mm, fontupper=\footnotesize, left=1mm, right=1mm, top=1mm, bottom=1mm]
\textbf{Instruction:} You are provided with historical OHLCV data of the past 20 days and a description of its statistics. Please predict how the stock will perform the next \texttt{<D>} day. Your response should be "Rise" or "Fall".

---

\textbf{Prompt: }
\texttt{<Embbedings of reprogrammed OHLCV>}

Statistics: The historical prices have a minimum close of \texttt{<min val>} \texttt{<min D>} days ago, a maximum close of \texttt{<max val>} \texttt{<max D>} days ago, and a median close of \texttt{<median val>} \texttt{<median D>} days ago. The overall trend is \texttt{<upward or downward>}...

Question: Given the reprogrammed OHLCV data and its statistics, how is the stock expected to perform in the next \texttt{<D>} days?
\end{tcolorbox}
\caption{Instruction and prompt for the Market Analyst.}
\label{fig:market_analyst_combined_prompt}
\end{figure}

\begin{figure*}[t]
\centering
\begin{tcolorbox}[colback=white, colframe=black, width=1\columnwidth, arc=0mm, outer arc=0mm, boxrule=0.5mm, fontupper=\footnotesize, left=1mm, right=1mm, top=1mm, bottom=1mm]
\textbf{Instruction:} You are provided with alpha factors derived from OHLCV data. Please predict the stock's movement based on the top contributing alpha factors. Your response should be "Rise" or "Fall".

---

\textbf{Prompt:} Alpha Factors: 
[Insert top contributing alpha factors here]

Descriptions: [Insert language descriptions for alpha factors]

Question: Based on the provided alpha factors, how is the stock expected to perform in the next \texttt{<D>} days?
\end{tcolorbox}
\caption{Instruction and prompt for the Alpha Expert.}
\label{fig:alpha_expert_combined_prompt}
\end{figure*}

\begin{figure*}[t]
\centering
\begin{tcolorbox}[colback=white, colframe=black, width=1\columnwidth, arc=0mm, outer arc=0mm, boxrule=0.5mm, fontupper=\footnotesize, left=1mm, right=1mm, top=1mm, bottom=1mm]
\textbf{Instruction:} You are provided with a summarized report of the stock's earnings call transcripts and fundamental metrics. Please predict whether the stock will rise or fall in the next quarter. Your response should include a prediction in one of the following five categories: "Strong Rise," "Moderate Rise," "No Change," "Moderate Fall," or "Strong Fall," followed by reasoning.

---

\textbf{Prompt:} Summarized Report: 
[Insert earnings call transcrpts and summarized fundamental metrics here]

Question: Based on the fundamental information, will the stock rise or fall in the next quarter?
\end{tcolorbox}
\caption{Instruction and prompt for the Fundamental Analyst.}
\label{fig:fundamental_analyst_combined_prompt}
\end{figure*}

\begin{figure*}[t]
\centering
\begin{tcolorbox}[colback=white, colframe=black, width=1\columnwidth, arc=0mm, outer arc=0mm, boxrule=0.5mm, fontupper=\footnotesize, left=1mm, right=1mm, top=1mm, bottom=1mm]
\textbf{Instruction:} You are provided with a summarized report of the stock. Please predict whether the stock will rise or fall in the next \texttt{<D>} day.

---

\textbf{Prompt:} Summarized Report: 
[Insert summarized report here]

Question: Based on the summarized report, will the stock rise or fall in the next \texttt{<D>} days?
\end{tcolorbox}
\caption{Instruction and prompt for the General Expert (Prediction Mode).}
\label{fig:general_expert_prediction_mode}
\end{figure*}

\begin{figure*}[t]
\centering
\begin{tcolorbox}[colback=white, colframe=black, width=1\columnwidth, arc=0mm, outer arc=0mm, boxrule=0.5mm, fontupper=\footnotesize, left=1mm, right=1mm, top=1mm, bottom=1mm]
\textbf{Instruction:} You are provided with summarized reports of two stocks. Please determine which stock will perform better in the next \texttt{<D>} day. Please output ``Stock AAA" or ``Stock BBB".

---

\textbf{Prompt:}

Summarized Report for Stock AAA: 
[Insert summarized report A here]

Summarized Report for Stock BBB:
[Insert summarized report B here]

Question: Based on the summarized reports, which stock will perform better in the next \texttt{<D>} days?
\end{tcolorbox}
\caption{Instruction and prompt for the General Expert (Ranking Mode).}
\label{fig:general_expert_ranking_mode}
\end{figure*}

\section{Examples of Datasets}

Figures \ref{fig:news_analyst_example}, \ref{fig:market_analyst_example}, \ref{fig:alpha_expert_example}, and \ref{fig:fundamental_analyst_example} illustrate the example datasets for the News Analyst, Market Analyst, Alpha Expert, and Fundamental Analyst LLMs, respectively, while Figure \ref{fig:general_expert_example} presents the example dataset for the General Expert LLM, which synthesizes the inputs from all expert LLMs.

\begin{figure*}
\centering

\begin{tcolorbox}[colback=white, colframe=black, width=1\columnwidth, arc=0mm, outer arc=0mm, boxrule=0.5mm, fontupper=\footnotesize, left=1mm, right=1mm, top=1mm, bottom=1mm]

\textbf{Instruction:} You are provided with a news article. Please predict how the stock will perform in the next \texttt{<D>} days. Your response should include your reasoning followed by a prediction of "Rise" or "Fall" in the specified format.

Format your response as follows: \\
Reasoning: [Your reasoning here] \\
Prediction: [Rise or Fall]

---

\textbf{Prompt:} \\
News Article: \\
Competition fears couldn’t rattle Apple stock this year, and 2024 looks bright, say Wedbush analysts.

And indeed, shares of the iPhone maker are up 49\% this year.

The Wedbush analysts, led by Dan Ives, wrote in a Thursday report, “The ’iPhone China demise narrative‘ was a great fictional story by the bears which is far from the reality as underlying mainland China growth remains strong for the December quarter based on our Asia supply chain checks and a key asset for the core iPhone franchise.”

Apple is still the firm’s “top tech pick,” thanks to “a strong iPhone 15 upgrade cycle,” an unproblematic holiday season leading into next year, in spite of competition from Huawei in China, and growth in the Services business. Wedbush rates Apple stock at Outperform with a \$250 price target. That’s 29\% upside from Thursday’s closing price of \$193.58.

Question: Given the information in the news article above, how is the stock expected to perform in the next 1 day?

---

\textbf{Model Response:} \\
Reasoning: The news article highlights strong performance and positive future prospects for Apple, particularly due to a strong iPhone 15 upgrade cycle and robust growth in the Services business. The positive outlook and the 29\% upside potential as per Wedbush analysts suggest investor confidence, leading to a potential rise in stock price. \\
Prediction: Rise

\end{tcolorbox}

\caption{Example data for the News Analyst LLM.}
\label{fig:news_analyst_example}

\end{figure*}

\begin{figure*}
    \centering
\begin{tcolorbox}[colback=white, colframe=black, width=1\columnwidth, arc=0mm, outer arc=0mm, boxrule=0.5mm, fontupper=\footnotesize, left=1mm, right=1mm, top=1mm, bottom=1mm]
\textbf{Historical OHLCV Data (Last 20 Days):}

\begin{tabular}{|c|c|c|c|c|c|}
\hline
\textbf{Date} & \textbf{Open} & \textbf{High} & \textbf{Low} & \textbf{Close} & \textbf{Volume} \\
\hline
2023-12-01 & 190.33 & 191.56 & 189.23 & 190.74 & 45,679,300 \\
2023-12-04 & 189.98 & 190.05 & 187.45 & 188.93 & 43,389,500 \\
2023-12-05 & 190.21 & 194.40 & 190.18 & 192.91 & 66,628,400 \\
2023-12-06 & 194.45 & 194.76 & 192.11 & 191.81 & 41,089,700 \\
2023-12-07 & 193.63 & 195.00 & 193.59 & 193.76 & 47,477,700 \\
2023-12-08 & 194.20 & 195.99 & 193.67 & 195.20 & 53,377,300 \\
2023-12-11 & 193.11 & 193.49 & 191.42 & 192.67 & 60,943,700 \\
2023-12-12 & 193.08 & 194.72 & 191.72 & 194.20 & 52,696,900 \\
2023-12-13 & 195.09 & 198.00 & 194.85 & 197.44 & 70,404,200 \\
2023-12-14 & 198.02 & 199.62 & 196.16 & 197.59 & 66,831,600 \\
2023-12-15 & 197.53 & 198.40 & 197.00 & 197.05 & 128,256,700 \\
2023-12-18 & 196.09 & 196.63 & 194.39 & 195.38 & 55,751,900 \\
2023-12-19 & 196.16 & 196.95 & 195.89 & 196.42 & 40,714,100 \\
2023-12-20 & 196.90 & 197.68 & 194.83 & 194.32 & 52,242,800 \\
2023-12-21 & 196.10 & 197.08 & 193.50 & 194.17 & 46,482,500 \\
2023-12-22 & 195.18 & 195.41 & 192.97 & 193.09 & 37,122,800 \\
2023-12-26 & 193.61 & 193.89 & 192.83 & 192.54 & 28,919,300 \\
2023-12-27 & 192.49 & 193.50 & 191.09 & 192.64 & 48,087,700 \\
2023-12-28 & 194.14 & 194.66 & 193.17 & 193.07 & 34,049,900 \\
2023-12-29 & 193.90 & 194.40 & 191.73 & 192.02 & 42,628,800 \\
\hline
\end{tabular}

\quad

\textbf{Instruction:} You are provided with historical OHLCV data of the past 20 days and a description of its statistics. Please predict how the stock will perform the next 1 day. Your response should be "Rise" or "Fall".

---

\textbf{Prompt:} \\
Embeddings of Reprogrammed OHLCV: [Insert embeddings here] \\

Statistics: The historical prices have a minimum close of \$188.93 on 19 days ago, a maximum close of \$197.59 on 10 days ago, and a median close of \$194.20 over the last 20 days. The overall trend appears to be slightly downward in the last 5 days.

Question: Given the reprogrammed OHLCV data and its statistics, how is the stock expected to perform in the next 1 day?

---

\textbf{Model Response:} \\
Prediction: Fall

\end{tcolorbox}

\caption{Example data for the Market Analyst LLM.}
\label{fig:market_analyst_example}

\end{figure*}

\begin{figure*}
    \centering
\begin{tcolorbox}[colback=white, colframe=black, width=1\columnwidth, arc=0mm, outer arc=0mm, boxrule=0.5mm, fontupper=\footnotesize, left=1mm, right=1mm, top=1mm, bottom=1mm]
\textbf{Top-5 Alpha Factors (Largest Values):}

\begin{tabular}{|c|l|}
\hline
\textbf{ID} & \textbf{Expression} \\
\hline
26 & 1 * (close - ts\_min(close, 30)) / (ts\_max(close, 30) - ts\_min(close, 30)) \\
5 & 1 * (close - vwap) / decay\_linear(rank(ts\_argmax(close, 30)), 2) \\
29 & 1 * ts\_rank(stddev(high, 10), 5) \\
27 & 1 * ts\_rank(correlation(rank(low), rank(adv15), 9), 15) \\
25 & 1 * ts\_rank(correlation(rank(high), rank(volume), 5), 5) \\
\hline
\end{tabular}

\quad

\textbf{Instruction:} You are provided with alpha factors derived from OHLCV data. Please predict the stock's movement based on the top contributing alpha factors. Your response should be "Rise" or "Fall."

---

\textbf{Prompt:} \\
Alpha Factors: \\
ID 26: 1 * (close - ts\_min(close, 30)) / (ts\_max(close, 30) - ts\_min(close, 30)) \\
ID 5: 1 * (close - vwap) / decay\_linear(rank(ts\_argmax(close, 30)), 2) \\
ID 29: 1 * ts\_rank(stddev(high, 10), 5) \\
ID 27: 1 * ts\_rank(correlation(rank(low), rank(adv15), 9), 15) \\
ID 25: 1 * ts\_rank(correlation(rank(high), rank(volume), 5), 5)

Descriptions: \\
- ID 26: Measures how close the current closing price is to the minimum and maximum close in the last 30 days. \\
- ID 5: Evaluates the deviation of the closing price from the VWAP, adjusted by a decayed rank. \\
- ID 29: Ranks the standard deviation of the high prices over the last 10 days. \\
- ID 27: Ranks the correlation between the low price and a 15-day average volume rank. \\
- ID 25: Ranks the correlation between the high price and volume over the last 5 days.

The comprehensive score derived from LightGBM is: 0.88 \\

Question: Based on the provided alpha factors, how is the stock expected to perform in the next 1 day?

---

\textbf{Model Response:} \\
Prediction: Rise

\end{tcolorbox}

\caption{Example data for the Alpha Expert LLM.}
\label{fig:alpha_expert_example}

\end{figure*}

\begin{figure*}

\begin{tcolorbox}[colback=white, colframe=black, width=1\columnwidth, arc=0mm, outer arc=0mm, boxrule=0.5mm, fontupper=\footnotesize, left=1mm, right=1mm, top=1mm, bottom=1mm]
\textbf{Summarized Earning Calls Transcript:}

Apple reported strong Q4 2023 financial results with total net sales of \$89.50 billion, a gross margin of \$40.43 billion, operating income of \$26.97 billion, and net income of \$22.96 billion. The company achieved a diluted earnings per share (EPS) of \$1.46. 

During the earnings call, Apple’s CEO Tim Cook highlighted the company's record performance in key markets, including India, China, and Latin America. iPhone revenue set a September quarter record at \$43.8 billion, up 3\% year-over-year, driven by strong demand in emerging markets. Apple's services segment also performed exceptionally well, setting an all-time revenue record of \$22.3 billion, a 16\% year-over-year increase. 

However, the Mac and iPad segments experienced declines in revenue due to challenging market conditions and tough comparisons to the prior year’s extraordinary results, where supply disruptions led to pent-up demand being fulfilled in the corresponding quarter. Despite these challenges, Apple continued to invest in future innovations, including the launch of new M3 chips and Apple Vision Pro.

Looking forward, Apple provided guidance that it expects iPhone revenue to grow year-over-year in the December quarter and anticipated strong gross margins of between 45\% and 46\%. The company also highlighted its strategic focus on emerging markets, with substantial growth opportunities in India.

\textbf{Fundamental Metrics:}

Total Net Sales: \$89.50 billion
Gross Margin: \$40.43 billion
Operating Income: \$26.97 billion
Net Income: \$22.96 billion
Earnings Per Share (EPS):
Basic EPS: \$1.47
Diluted EPS: \$1.46

---

\textbf{Instruction:} You are provided with a summarized report of the stock's earnings call transcript and fundamental metrics. Please predict whether the stock will rise or fall in the next quarter. Your response should include a prediction in one of the following five categories: "Strong Rise," "Moderate Rise," "No Change," "Moderate Fall," or "Strong Fall," followed by reasoning.

---

\textbf{Prompt:} \\
Summarized Report: Apple's Q4 2023 earnings report shows strong financial performance with a total net sales of \$89.50 billion, a gross margin of \$40.43 billion, operating income of \$26.97 billion, and net income of \$22.96 billion. The earnings per share (EPS) are \$1.46 (diluted). 

During the earnings call, CEO Tim Cook emphasized record iPhone revenue in several key markets and a strong performance in the services segment. The company expects continued growth in iPhone revenue and strong gross margins in the upcoming quarter, supported by ongoing investments in innovation and expansion into emerging markets.

Question: Based on the summarized report, will the stock rise or fall in the next quarter?

---

\textbf{Model Response:} \\
Reasoning: Apple's robust financial metrics, particularly the record iPhone and services revenue, coupled with the positive outlook for the December quarter, suggest strong market confidence. The company's strategic focus on emerging markets, along with its commitment to innovation, further supports this positive sentiment. Given these factors, it is likely that the stock will experience a moderate rise in the next quarter.

Prediction: Moderate Rise

\end{tcolorbox}

\caption{Example data for the Fundamental Analyst LLM.}
\label{fig:fundamental_analyst_example}

\end{figure*}

\begin{figure}
\centering
\begin{tcolorbox}[colback=white, colframe=black, width=1.0\columnwidth, arc=0mm, outer arc=0mm, boxrule=0.5mm, fontupper=\footnotesize, left=1mm, right=1mm, top=1mm, bottom=1mm]

\textbf{Input from News Analyst:}
\begin{itemize}
    \item \textbf{Reasoning:} The news article highlights strong performance and positive future prospects for Apple, particularly due to a strong iPhone 15 upgrade cycle and robust growth in the Services business. The positive outlook and the 29\% upside potential as per Wedbush analysts suggest investor confidence, leading to a potential rise in stock price.
    \item \textbf{Prediction:} Rise
\end{itemize}

\textbf{Input from Market Analyst:}
\begin{itemize}
    \item \textbf{Historical OHLCV Data:} The stock has shown a slightly downward trend over the last 5 days, with a recent close of \$192.02.
    \item \textbf{Prediction:} Fall
\end{itemize}

\textbf{Input from Alpha Expert:}
\begin{itemize}
    \item \textbf{Top-5 Alpha Factors:}
    \begin{itemize}
        \item ID 26: 1 * (close - ts\_min(close, 30)) / (ts\_max(close, 30) - ts\_min(close, 30))
        \item ID 5: 1 * (close - vwap) / decay\_linear(rank(ts\_argmax(close, 30)), 2)
        \item ID 29: 1 * ts\_rank(stddev(high, 10), 5)
        \item ID 27: 1 * ts\_rank(correlation(rank(low), rank(adv15), 9), 15)
        \item ID 25: 1 * ts\_rank(correlation(rank(high), rank(volume), 5), 5)
    \end{itemize}
    \item \textbf{Prediction:} Rise
\end{itemize}

\textbf{Input from Fundamental Analyst:}
\begin{itemize}
    \item \textbf{Reasoning:} Apple's robust financial metrics, particularly the record iPhone and services revenue, coupled with the positive outlook for the December quarter, suggest strong market confidence. The company's strategic focus on emerging markets, along with its commitment to innovation, further supports this positive sentiment.
    \item \textbf{Prediction:} Moderate Rise
\end{itemize}

---

\textbf{Instruction:} You are provided with a summarized report of the stock. Please predict whether the stock will rise or fall the next 1 day.

Format your response as follows: Reasoning: [Your reasoning here] Prediction: [Rise or Fall].

---

\textbf{Model Response:} \\
Reasoning: While the Market Analyst predicts a fall based on recent downward trends in the OHLCV data, the overall sentiment from the News Analyst, Alpha Expert, and Fundamental Analyst is positive. The strong financial metrics, record iPhone and services revenue, and positive forward guidance from the earnings call provide a solid foundation for a potential rise. The Alpha Expert’s positive alpha factors also support this outlook. Therefore, despite the short-term downward trend, the overall stock movement is expected to be upward.

Prediction: Rise

\end{tcolorbox}

\caption{Example data for the General Expert LLM.}
\label{fig:general_expert_example}

\end{figure}

\end{document}